\documentclass{article}

 \PassOptionsToPackage{numbers, compress}{natbib}
 \usepackage[final]{nips_2016}

\usepackage[utf8]{inputenc} % allow utf-8 input
\usepackage[T1]{fontenc}    % use 8-bit T1 fonts
\usepackage{hyperref}       % hyperlinks
\usepackage{url}            % simple URL typesetting
\usepackage{booktabs}       % professional-quality tables
\usepackage{amsfonts}       % blackboard math symbols
\usepackage{nicefrac}       % compact symbols for 1/2, etc.
\usepackage{microtype}      % microtypography

\usepackage{graphicx}
\usepackage[usenames,dvipsnames]{xcolor}

\usepackage{amssymb}
\usepackage{amsmath}
\usepackage{amsthm}

\newcommand\transpose{^{\top}}

\newcommand\bgamma{ { \boldsymbol{\gamma} } }
\newcommand\hgamma{ { \hat{\gamma} } }
\newcommand\hbgamma{ \widehat{ \boldsymbol{\gamma } } }

\newcommand\hbGamma{ \widehat{ \boldsymbol{\Gamma} } }

\newcommand\bbE{\mathbb{ E } }

\newcommand\hbV{\widehat{ \mathbf{V} } }

\newcommand \hbS{\widehat{ \mathbf{S}}}
\newcommand \bS{\mathbf{S}}

\newcommand \hG{\widehat{G}}

\newcommand \hF{\widehat{F}}

\newcommand \hbv{\widehat{ \mathbf{v}}}

\newcommand \hbC{\widehat{ \mathbf{C}}}

\newcommand \bC{{ \mathbf{C}}}

\newcommand \bX{{ \mathbf{X}}}
\newcommand \bx{{ \mathbf{x}}}
\newcommand \hbx{\widehat{ \mathbf{x}}}

\DeclareMathOperator*{\argmin}{arg\,min}

\usepackage{environ}
\makeatletter
\NewEnviron{Lalign}{\tagsleft@true\begin{align}\BODY\end{align}}
\makeatother

\makeatletter
\renewcommand\NAT@bibsetnum[1]{\settowidth\labelwidth{\@biblabel{#1}}%
   \setlength{\leftmargin}{\bibindent}\addtolength{\leftmargin}{\dimexpr\labelwidth+\labelsep\relax}%
   \setlength{\itemindent}{-\bibindent}%
   \setlength{\listparindent}{\itemindent}
\setlength{\itemsep}{\bibsep}\setlength{\parsep}{\z@}%
   \ifNAT@openbib
     \addtolength{\leftmargin}{\bibindent}%
     \setlength{\itemindent}{-\bibindent}%
     \setlength{\listparindent}{\itemindent}%
     \setlength{\parsep}{0pt}%
   \fi
}
\makeatother

\newlength{\tablerulesep}
\title{Cross-validation based Nonlinear Shrinkage}

\author{
  Daniel Bartz\thanks{Use footnote for providing further
    information about author (webpage, alternative
    address)---\emph{not} for acknowledging funding agencies.} \\
  emetriq GmbH, Hamburg\\
  \texttt{d.bartz@emetriq.com} \\
}

\begin{document}

\maketitle

% % % % % % % % % % % % % % % % % % abstract
% % % % % % % % % % % % % % % % % % abstract
% % % % % % % % % % % % % % % % % % abstract
\begin{abstract}
Many machine learning algorithms require precise estimates of covariance matrices. The sample covariance matrix performs poorly in high-dimensional settings, which has stimulated the development of alternative methods, the majority based on factor models and shrinkage. Recent work of Ledoit and Wolf has extended the shrinkage framework to Nonlinear Shrinkage (NLS), a more powerful covariance estimator based on Random Matrix Theory.

Our contribution shows that, contrary to claims in the literature, cross-validation based covariance matrix  estimation (CVC) yields comparable performance at strongly reduced complexity and runtime. On two real world data sets, we show that the CVC estimator yields superior results than competing shrinkage and factor based methods.
\end{abstract}

\section{Introduction}
Covariance matrices are an integral part of many algorithms in signal processing, machine learning and statistics. 
The standard estimator is the sample covariance matrix $\mathbf{S}$. It has excellent properties in the limit of large sample sizes $n$: its entries are unbiased and consistent. %\citep{HasTibFri08}.  
For sample sizes  of the order of the dimensionality $p$ or even smaller, the estimation quality degrades, its entries have a high variance and the spectrum has a large systematic error. In particular, large eigenvalues are overestimated and small eigenvalues  underestimated, the condition number is large and the matrix difficult to invert \citep{MarPas67,EdeRao05}.

Shrinkage of  $\mathbf{S}$ towards a biased estimator with lower variance,
\begin{align}\mathbf{C}^{sh} := (1
-\lambda) \mathbf{S} + \lambda \mathbf{T},
\end{align}
improves performance in high-dimensional settings and  is  governed by a single regularization parameter \citep{Ste56}.
A common choice for the shrinkage target is  $\mathbf{T} = p^{-1} \text{trace}(\mathbf{S}) \mathbf{I}$. Ledoit and Wolf proposed \emph{LW-Shrinkage}, a  fast, easy to implement and numerically robust method to estimate the optimal shrinkage intensity $\lambda$  \cite{LedWol04} .

The downside of shrinkage is limited flexibility: shrinkage is not flexible enough to improve estimation in the low- and high-variance regions of the spectrum at the same time.
In 1975, Stein already proposed a first algorithm which corrects each eigenvalue separately.
Still, \emph{spectrum correction} has found little application in practice, as the estimation of the large number of correction factors  suffers from  high variance.

The past few years have shown increased research  on the Random Matrix Theory  (RMT) of covariance matrix estimation. The standard problem of RMT is the derivation of the distribution of the sample eigenvalues from the distribution of the population eigenvalues. The relationship is governed by the Mar\v{c}enko-Pastur law. In covariance matrix estimation the direction of inference is inverted: the sample eigenvalues are given and we are interested in the population covariance matrix. 
\cite{Kar08} and \cite{ledoit2011eigenvectors} discuss covariance estimation by inversion of the Mar\v{c}enko-Pastur law, research which led to the state-of-the-art approach  \emph{Nonlinear Shrinkage (NLS)} \cite{ledoit2012nonlinear}.

In this article, we propose CVC, a cross-validation-based spectrum correction approach which estimates the corrected eigenvalues on hold-out sets. The CV-approach optimizes the same distance measure as NLS and has the advantage of conceptual simplicity, smaller runtime and higher numerical stability. Theoretical guarantees for cross-validation require hypothesis or error stability of the algorithm under consideration \citep{vapnik1982estimation,devroye2013probabilistic,kearns1999algorithmic}. The algorithmic instability of the eigenvalue decomposition, which we  illustrate in detail, makes it very hard to obtain theoretical guarantees for CVC, hence this is  left for future work.
This lack of a theoretical foundation is compensated by very strong empirical results: the evaluation of CVC on real world data sets from two domains shows that CVC is the best performing general purpose covariance estimator. 

\section{Improved estimation by spectrum correction}
\label{sec:rmt:others}
The sample covariance matrix is a symmetric matrix. It can be decomposed as
\begin{align*}
\hbS = \hbV \, \hbGamma \, \hbV \transpose = {\sum}_i \hgamma_i \hbv_i \hbv_i \transpose,
\end{align*}
where $\hbV$ is a rotation matrix given by the eigenvectors $\hbv_i$  and $\hbGamma$ is the diagonal matrix of sample eigenvalues~$\hgamma_i$.
Shrinkage to the standard target \cite{LedWol04} can also be seen as spectrum correction:
\begin{align}
\label{eq:rmt:eigenvalue_shrinkage}
\hgamma_i^\mathrm{shr} := (1 - \lambda) \hgamma_i + \lambda p^{-1} {\sum}_i \hgamma_i.
\end{align}
The convex combination %in eq.\,\eqref{eq:rmt:eigenvalue_shrinkage} 
is governed by a single parameter. More flexible approaches modify each eigenvalue individually. The first one was proposed by Stein in his 1975 Rietz lecture, the mathematical formulation, extensions and alternative approaches can be found in \cite{stein1986lectures,dey1985estimation}.

%The obtained eigenvalues may no longer decrease monotonically. Hence, it is recommended to apply \emph{isotonic regression} which yields a monotonically decreasing sequence (for details, see Section~\ref{sec:rmt:cv_performance})
Spectrum correction has found little application in practice. Although the estimation of $p$ correction factors is very powerful, the increased flexibility comes at the cost of increased estimation errors. \citep{LedWol04} show that Shrinkage with a single regularization parameter tends to yield better performance. 

\section{Random Matrix Theory (RMT) for spectrum correction}
\label{sec:rmt:rmt}
The field of RMT studies the theoretical properties of large random matrices \cite{EdeRao05}. The majority of  the research focuses on the task of inferring the distribution of the sample eigenvalues from the distribution of the entries in a random matrix. 
For covariance matrices, this distribution is given by the Mar\v{c}enko-Pastur law and its generalizations \citep{MarPas67,silverstein1995strong}.
Consequently, the first applications of RMT to covariance estimation only used the sample eigenvalue distribution to estimate the number of factors in a factor model \cite{Ros02,Lal00}.
% to test the hypothesis of all population eigenvalues being identical. For each sample eigenvalue 
% If this hypothesis had be rejected, a factor model would incorporarte a factor for the direction.
%\cite{Ros02} set the number of factors  to the number of sample eigenvalues larger than the largest eigenvalue of a random matrix, assuming that the corresponding eigendirections reflect some real structure. Their model is equivalent to a PCA factor model based  on the correlation matrix.
%\cite{} propose a very similar approach based on a probabilistic PCA model.

%\subsection{Inverting the Mar\v{c}enko-Pastur law}
%\label{sec:rmt:inversion}
El Karoui \cite{Kar08} made the first real attempt to harness the possibilities of RMT. He numerically inverts the Mar\v{c}enko-Pastur law, which describes the relationship between the distributions of population and sample eigenvalues,  in order to estimate the population eigenvalues from the sample eigenvalues.
%The following paragraphs give a very short introduction to this approach.

\paragraph{Stieltjes transforms and the MP-law}
RMT is naturally formulated in terms of distributions: the population eigenvalues are not fixed, but drawn from a distribution. The distribution of the sample eigenvalues then depends on the distribution of population eigenvalues.

\cite{Kar08} considers \emph{asymptotics at fixed spectral distribution}: the cumulative distribution function (cdf) of the population eigenvalues $H_p$ is kept constant: $ H_p = H_\infty, \,  \forall \, p$. Then the empirical cdf $\hF_p$ of the sample eigenvalues is asymptotically non-random and the relation between $H_\infty$ and $F_\infty$ is given by the Mar\v{c}enko-Pastur law.

Many derivations and proofs in RMT rely on the Stieltjes transform of  the distribution of sample eigenvalues. This often simplifies the theoretical analysis.
The Stieltjes transform of the cdf $\hF_p$ on $\mathbb R$ is a function with complex arguments given by
\begin{align}
\label{eq:rmt:stieltjes}
m_{\hF_p}(z) := \int \frac{d\hF_p(x)}{x-z},
\quad
\forall z \in \mathbb C^+, \qquad  \mathbb C^+ := \{ z \in \mathbb C : \mathrm{Im}(z) > 0 \}
\end{align}

\cite{Kar08} analyzes $\hG_p$, the empirical spectral distribution of the $n\times n$-matrix $n^{-1} \bX \transpose \bX$. In Machine Learning, this is called the linear kernel matrix. 
%\citep{scholkopf2002learning}. 
Its eigenvalues are, excluding zero eigenvalues, equal to the sample eigenvalues.

\cite{silverstein1995strong} derived the most general form of the  Mar\v{c}enko-Pastur law, which \cite{Kar08} uses to relate (the Stieltjes transform of) $\hG_p$ to the unknown distribution of population eigenvalues $H_\infty$:
\begin{align}
\label{eq:rmt:Silverstein}
m_{G_\infty} (z) = - \left( z - c \int \frac{\gamma dH_\infty(\gamma)}{1 + \gamma m_{G_\infty}(z)}  \right)^{-1}, \qquad c := \frac{n}{p}.
\end{align}
%Note that the Stieltjes transform $m_{G_\infty} (z)$ appears on both sides of the equation.

\paragraph{Solving for the distribution of population eigenvalues} To estimate the population eigenvalues, one has to find the $H_\infty$ which best satisfies eq.~\eqref{eq:rmt:Silverstein} for $m_{\hG_p}$, the Stieltjes transform of $\hG_p$. 
To make this feasible, 
 \cite{Kar08} makes two approximations: 
(i)
he approximates $dH_\infty(\gamma)$, the pdf of population eigenvalues, by a sum of delta peaks:
\begin{align}
\label{eq:delta_approx}
dH_\infty(\gamma) 
\approx 
{\sum}_{k=1}^K w_k \delta (t_k - \gamma ), 
\qquad
w_k \geq 0 
\;
\mathrm{and} 
\;
{\sum}_k w_k = 1,
\end{align}
where the $t_k$ form a predefined grid. It is possible to use different or include additional basis functions.
(ii)
Instead of optimizing eq.\,\eqref{eq:rmt:Silverstein} for all $z \in \mathbb C^+$, he only considers a discrete set of $J$ points  $(z_j , m_{G_p}(z_j))$.
% He does not optimize eq.\,\eqref{eq:rmt:Silverstein} for all $z \in \mathbb C^+$. Instead, he only optimizes $H_\infty$ for a set of $J$ points  $(z_j , m_{G_p}(z_j))$.

Plugging the approximation eq.\,\eqref{eq:delta_approx} into the MP-law eq.\,\eqref{eq:rmt:Silverstein}
 leads to a set of $J$  complex errors  
 %$e_j$ given by
\begin{align*}
e_j
:=
\frac{1}{m_{\hG_p}(z_j)}  + z_j + \frac{p}{n} \sum_{k=1}^K w_k \frac{t_k}{1 + m_{\hG_p}(z_j) t_k},
\end{align*}
which can be minimized by searching over the weights $w_k$. In the next step, the $p$ population eigenvalues are obtained by calculating  the quantiles of $dH_\infty(\gamma) $.
%
%
%
%\begin{align}
%- \frac{1}{v_\infty(z)} = z - c \int \frac{\gamma dH_\infty}{1 + \gamma v_\infty(z)}
%\end{align}

Unfortunately, El Karoui's approach has a set of drawbacks:
(i)
% \item 
% The inversion is not unique and therefore a  prior or parametric ansatz has to be applied. 
%
the largest eigenvalue of the covariance matrix is often isolated from the bulk. The discrete approximation of the population spectrum makes this case problematic.
(ii)
It is not  clear how the choices of the set $(z_j , m_{\hG_p}(z_j))$ and the grid $t_k$ affect the algorithm.
(iii)
The description of the algorithm is not clear. \cite{ledoit2012nonlinear} state that they were not able to reproduce the results and that the original code is not available.
(iv)
Most importantly, for spectrum correction, \emph{the population eigenvalues are not optimal} w.r.t.\ the  expected squared error (ESE), because the sample eigenvectors differ from the population eigenvectors. 
 Instead, the optimal corrected eigenvalues are given by the (spectrum) oracle  $\bgamma^\star$.
\begin{align}
\label{eq:rmt:optimal_spectrum}
\bgamma^\star 
:= \argmin_\gamma \left\|  \bC - {\sum}_i \gamma_i \hbv_i \hbv_i \transpose \right\|^2
\qquad \Longleftrightarrow  \qquad 
\gamma^\star_i 
= \hbv_i \transpose \bC \hbv_i.
\end{align}
%In the following,  $\bgamma^\star$ is called 
%\end{itemize}

These four  aspects make it difficult to apply the approach of El Karoui in practice: no publication reports a successful  application to a real world data set.
% portfolio simulations exists in which a competetive performance was achieved.
%

\subsection{Adapting RMT to the  covariance estimation problem}
\label{sec:rmt:adapting}
%
%In the last subsection,  it was  mentioned that estimating the population eigenvalues is not optimal for spectrum correction. 
%
Recently, \cite{ledoit2011eigenvectors} quantified the relationship between sample and population eigenvectors. Based on this relationship, they derived the distribution of the variances in the direction of the sample eigenvectors in eq.\,\eqref{eq:rmt:optimal_spectrum}. 
This directly yields an estimator for the spectrum oracle $\bgamma^\star$ which \cite{ledoit2012nonlinear} call \emph{Nonlinear Shrinkage (NLS)}:
% In other words: Given the population eigenvalues, their result allows to calculate the desired variances in the sample eigendirection:
\begin{align}
\label{eq:rmt:NLS_speccorr}
\hgamma_i^\mathrm{NLS} 
% = \hgamma_i \left| 1 - \frac{p}{n} -\frac{p}{n} \hgamma_i \breve{m}^\star_F (\hgamma_i) \right|^{-2},
:= \hgamma_i \left| 1 - c^{-1} - c^{-1} \hgamma_i \hat{m}_{F_\infty} (\hgamma_i) \right|^{-2},
\end{align}
where $\hat{m}_{F_\infty}(z)$ is an estimate of the Stieltjes transform of the limit distribution of the sample eigenvalues\footnote{The Stieltjes transform $\hat{m}_{F_\infty} (z)$ is only defined for $z$ with positive imaginary parts (see eq.\,\eqref{eq:rmt:stieltjes}). For real values, one defines $\hat{m}_{F_\infty} (\hgamma_i) := \lim_{y \rightarrow 0^+}\hat{m}_{F_\infty} (\hgamma_i + i y)$.   }.

\cite{ledoit2012nonlinear} obtain $\hat{m}_{F_\infty}(z)$ from the  Mar\v{c}enko-Pastur law and an estimate of the population eigenvalues. 
They propose an alternative to the inversion procedure presented in the last section which directly optimizes for the population eigenvalues.

The simulations in the next section use  a different algorithm from the same authors which is more stable\footnote{Personal correspondence with Olivier Ledoit.} \citep{ledoit2014spectrum}.
 % and
%The last section showed an algorithm for the estimation of the  population eigenvalues.
%To recap, spectrum correction by eq.\,\eqref{eq:rmt:NLS_speccorr} consists of two steps:
%\begin{enumerate}
%%
%\item Optimize for the population eigenvalues which have generated the sample eigenvalues.
%%
%\item Use the estimate of the population eigenvalues to calculate the desired variances.
%\end{enumerate}
% The main problem is the estimation of the population eigenvalues, as discussed in the last section. 
% \cite{ledoit2012nonlinear} propose an alternative inversion procedure which directly optimizes for the population eigenvalues and
% \cite{ledoit2014spectrum} present an improved version  which is more stable\footnote{personal correspondence}. 
It is based on an alternative formulation of eq.\,\eqref{eq:rmt:Silverstein} in  \citep{silverstein1995strong}:
 for the limit distribution of sample eigenvalues $F_\infty(z)$, the MP-law is given by
\begin{align}
\label{eq:rmt:SilversteinII}
m_{F_\infty} (z) =  \int \frac{ dH_\infty(\gamma)}{\gamma (1 - c^{-1} - c^{-1} z m_{F_\infty}(z)) - z} .
\end{align}
\cite{ledoit2014spectrum} discretize eq.\,\eqref{eq:rmt:SilversteinII} and derive a function $Q : [0,\infty)^p \rightarrow [0,\infty)^p$ which maps population eigenvalues to sample eigenvalues. 
The function $Q$ is used to 
formulate a (non-convex) optimization problem for the population eigenvalues, solvable by Sequential Linear Programming (SLP):
\begin{align*}
\hbgamma^{\mathrm{pop.}} := \argmin_\bgamma \| Q (\bgamma) - \hbgamma  \|^2.
\end{align*}

% % % % % % % % % % % % % % % % % % % % % % % % % % % % % % % % % % % % % % % % % % % % % % % % % % % % % % % % % % % % % % % 
\section{Using cross-validation for covariance estimation}
\label{sec:rmt:cv}
\cite{ledoit2012nonlinear} state that anonymous reviewers proposed to use leave-one-out (loo) cross-validation to estimate the variances in eq.\,\eqref{eq:rmt:optimal_spectrum}. The \emph{loo-cross-validation covariance (CVC)} estimator is defined by
\begin{align}
\label{eq:rmt:loocv}
 \hgamma_i^\mathrm{loo-CVC} 
 := n^{-1} {\sum}_{t=1}^n \left( \hbv^{\mathrm{loo}(t)}_i {} \transpose \bx_t \right)^2,
 \end{align}
where $\hbv^{\mathrm{loo}(t)}_i$ is the $i^{th}$ sample eigenvector of the sample covariance $\hbS^{\mathrm{loo}(t)}$, for which the $t^{th}$ data point has been removed from the data set. The 
vectors $\hbv^{\mathrm{loo}(t)}_i$ and $\bx_t$ are independent, this makes $\hgamma_i^\mathrm{loo-CVC}$ an unbiased estimator\footnote{Note that the normalization by $n$ only yields an unbiased estimate if the data is assumed to have zero mean.} of the variance in the directions $\hbv^{\mathrm{loo}(t)}_i$:
\begin{align}
\label{eq:rmt:loocv_exp}
\bbE \left[ \hgamma_i^\mathrm{loo-CVC} \right] 
& =  n^{-1} {\sum}_{t=1}^n   \bbE \left[  \hbv^{\mathrm{loo}(t)}_i {}\transpose   \bx_t \bx_t   \hbv^{\mathrm{loo}(t)}_i \right] 
% \\
% \notag
 =     \bbE \left[  \hbv^{\mathrm{loo}(t)}_i {}\transpose  \bC   \hbv^{\mathrm{loo}(t)}_i \right].
 \end{align}
\cite{ledoit2012nonlinear} state that the CVC approach has the advantage of conceptual simplicity, while NLS has three clear advantages: 
(i)
NLS can estimate any smooth function of the eigenvalues. In particular, it is possible to directly optimize for the precision matrix.
(ii)
NLS yields clearly smaller estimation error than CVC.
(iii) 
NLS is faster  than the CVC approach. CVC requires $n+1$  eigendecompositions of $p$-dimensional sample covariance matrices. For large $p$, this is very time consuming.
The following paragraphs  discuss these aspects. 
% Before I start, note that there is a clear advantage of CVC:  NLS relies on the  Mar\v{c}enko-Pastur law and thereby assumes Gaussianity. CVC, on the other hand, is less dependent on distributional assumptions.

% \paragraph{CVC for $n < p$}
% If the dimensionality exceeds the number of observations, it is possible to speed up CVC by calculating the eigenvalue deco

% % % % % % % % % % % % % % % % % % % % % % % % % % % % % % % % % % % % % % % % % % % % % % % % % % % % 
\paragraph{Precision matrix estimation}
Intriguingly, the optimal spectrum correction is different for covariance and precision matrix:
%\footnote{\cite{ledoit2013optimal} also discuss optimal estimation under Stein's Loss. 
% It is not clear under which circumstances Stein's Loss is superior to ESE or ESE of the precision matrix.}:
\begin{align}
\label{eq:rmt:optimal_spectrum_precision}
\bgamma^\diamond 
:= {\argmin}_\gamma \left\|  \bC^{-1} - {\sum}_i \gamma_i^{-1} \hbv_i \hbv_i \transpose \right\|^2 
\qquad \Longleftrightarrow  \qquad 
{\gamma^\diamond_i}^{-1}
= \hbv_i \transpose \bC^{-1} \hbv_i 
\neq {\gamma_i^\star}^{-1}.
\end{align}
In the following, $\bgamma^\diamond$ is called the precision oracle.
Based on the results on the relationship between sample and population eigenvectors, \cite{ledoit2011eigenvectors} derive an estimator for $\bgamma^\diamond$:
\begin{align*}
%\label{eq:rmt:precisioncorr}
\hgamma_i^\mathrm{NLS-precision} 
:= \hgamma_i \left( 1 - c^{-1} - 2 c^{-1} \hgamma_i \mathrm{Re} [ \hat{m}_{F_\infty} (\hgamma_i) ] \right)^{-1}.
\end{align*}
Many algorithms require estimates of the precision matrix and hence it seems to be a good idea to use a covariance estimator with minimal ESE for the precision.
% estimate directly optimize for the precision matrix. 
One has to keep in mind that in a practical application, both definitions of estimation error only serve as  proxies for the performance. It is not possible to say in general that one of the two will work better. 

The main advantage of spectrum correction compared to standard Shrinkage, the individual optimization of eigenvalues, holds independently of the specific loss function: spectrum correction is flexible  enough to precisely estimate low- and high-variance directions, while in LW-Shrinkage,  strong directions dominate the loss function and large relative estimation errors in weak directions are neglected.
%\end{itemize}
%My hypothesis is that  most applications will not display a pronounced performance difference between $\hat \gamma^\star$ and $\hat \gamma^\diamond$. Appendix~\ref{ch:systematiclda} provides simulations which show that for LDA this is indeed the case.
%
The supplemental material provides simulations which compare LDA  classification performances for
$ \gamma^\star$ and $ \gamma^\diamond$.
The results support the hypothesis that  most applications will not display a pronounced performance difference between both estimators.

% % % % % % % % % % % % % % % % % % % % % % % % % % % % % % % % % % % % % % % % % % % % % % % % % % % % 
\paragraph{Estimation error of the CVC approach}
\label{sec:rmt:cv_performance}
\cite{ledoit2012nonlinear} show that loo-CVC  does not yield competitive results. In the following paragraphs, we (i)~explain what makes the theoretical analysis difficult, (ii)~illustrate why loo-CVC is not competitive and (iii)~propose \emph{iso-loo-CVC}, an improved  CVC estimator.

% % % % % % % % % % % % % % % % % % % % % % 
%\paragraph{Theoretical analysis of loo-CVC} 
The loo-CVC estimator yields unbiased estimates of the variance in the eigendirections for $n-1$ data points, as shown in eq.\,\eqref{eq:rmt:loocv_exp}, and is therefore \emph{asymptotically unbiased}.
In general, \emph{convergence} and the \emph{convergence rate} of cross-validation depend on the algorithm under consideration \citep{vapnik1982estimation,devroye2013probabilistic} and require some notion of \emph {algorithmic stability} \citep{kearns1999algorithmic}: removing a single data point from the data set is assumed to have negligible influence on the obtained solution (hypothesis stability) or out-of-sample-error (error stability). The eigendecomposition is not algorithmically stable: the removal of a single data point can completely change the obtained eigendirections. In fact, the dependency between the loo-eigendirections $\hbv^{\mathrm{loo}(t)}_i$ causes a high estimation error in 
the loo-estimates of the variances (eq.\,\eqref{eq:rmt:loocv}). A detailed theoretical analysis is left for future research.
 
%because the eigendecomposition suffers from algorithmic instability .
% do not yield low estimation error because of the dependency of the $\hbv^{\mathrm{loo}(t)}_i$. 

% % % % % % % % % % % % % % % % % % % % % % 
%\paragraph{Illustration of the loo-CVC estimation error} 
\begin{figure} 
\begin{center}
\includegraphics[width= 0.8 \linewidth]{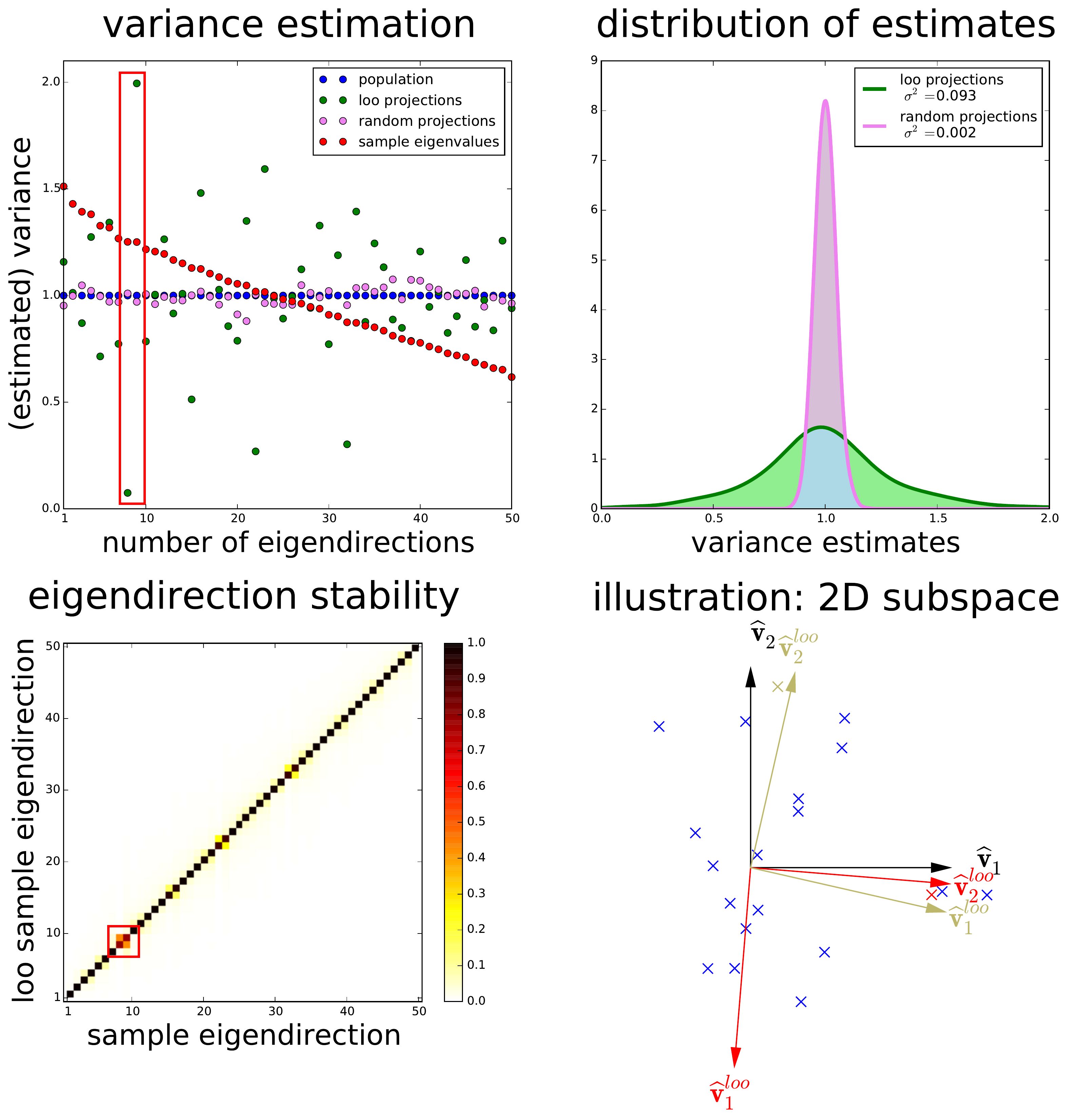}
\caption[Illustration of the estimation error of loo-CVC.]{Illustration of the estimation error of loo-CVC.}
\label{fig:loo_failure}
  % generated by sotm_paper_whitening.m
\end{center}
\end{figure}
Figure~\ref{fig:loo_failure} illustrates the high estimation error of loo-CVC for Gaussian data with an identity population covariance ($p=50$, $n=1000$). For identity covariance, the loo-sample eigenvectors are random projections which are dependent over folds. 
To show the effect of this dependence, loo-sample eigenvectors are compared to projections which are entirely random.
\\
The upper left plot shows (estimated) variances given by population and sample eigenvalues as well as estimates based on loo-CVC and random projection.
It can be seen that the variance of the  estimates based on loo-CVC estimates is huge compared to those based on random projections. 
The upper right plot shows the corresponding  distributions of the variance estimates from loo-CVC and random projections (based on $R=100$ repetitions).  The variances of the random projections follow a  $\chi^2_n$ distribution  (normalized by $n$) and hence have variance $2 n^{-1}$. The variance of the loo-projections is far higher.
\\
To explain this effect, the lower left plot shows the stability of the eigenvectors in the upper left plot, measured by the scalar product of sample eigenvectors and loo-sample eigenvectors (averaged over folds).
One can see that there is a high instability of eigenvectors 8 and 9.  The upper left plot shows that this is exactly where (i) the sample eigenvalues are very similar and (ii) the loo-CVC estimates deviate strongly from the population. To be precise, the loo-CVC corrections for the 8$^{th}$ (larger) and 9$^{th}$ (smaller)  sample eigenvalues are under- and overestimated, respectively. 
\\
\definecolor{DarkKhaki}{HTML}{BDB76B}
The plot to the lower right illustrates why this happens. It shows a subspace where both sample eigendirections $\hbv_1$ and $\hbv_2$ have very similar sample eigenvalues. Removing the yellow datapoint {\color{DarkKhaki} $\times$}, with a high projection on $\hbv_2$, does not considerably change the eigendirections. We have
\begin{align*}
| \hbv_1\transpose  {\color{DarkKhaki} \hbx} |
< | \hbv_2\transpose  {\color{DarkKhaki} \hbx} |
\qquad \Rightarrow \qquad 
| {\color{DarkKhaki}  \hbv^{\mathrm{loo}}_1} \transpose  {\color{DarkKhaki} \hbx} |
< | {\color{DarkKhaki}  \hbv^{\mathrm{loo}}_2} \transpose  {\color{DarkKhaki} \hbx} |.
\intertext{Removing the red data point {\color{Red} $\times$} with a high projection on $\hbv_1$, on the other hand, changes the eigendirections completely: now {\color{Red} $\hbv^{\mathrm{loo}}_2$ } points in the direction of  data point {\color{Red} $\times$} and we have
}
| \hbv_2\transpose  {\color{Red} \hbx} |
< | \hbv_1\transpose  {\color{Red} \hbx} |
\qquad \Rightarrow \qquad 
| {\color{Red}  \hbv^{\mathrm{loo}}_1}  \transpose  {\color{Red} \hbx} |
< |{\color{Red}  \hbv^{\mathrm{loo}}_2}  \transpose  {\color{Red} \hbx} |.
\end{align*}
It can be seen that no matter  on which sample eigendirections the projection is higher, the projection on $\hbv^{\mathrm{loo}}_2$ is always higher than on $\hbv^{\mathrm{loo}}_1$. Therefore, the  weaker direction has  a strongly increased loo-CVC variance estimate.
\\
A straightforward fix for this behavior is replacing loo-cross-validation by K-fold cross-validation. Using K-fold CV, it is highly unlikely that the entire set of hold-out data points has a high projection on the same eigendirection and is hence affected in the same way. This is shown in simulations after introducing isotonic regression.

% \paragraph{Reducing estimation error by isotonic regression} 
loo-CVC yields eigenvalues which are not ordered. There is no reason to assume that a sample eigendirection with larger eigenvalue should have lower variance than one with smaller eigenvalue. We propose to apply  \emph{isotonic regression} \citep{barlow1972statistical}: isotonic regression searches for the decreasing sequence with minimum squared distance to the estimates:
%\begin{align}
%\label{eq:rmt:iso-loocv}
%( \hgamma_1^\mathrm{iso-loo-CVC} , \dots, \hgamma_p^\mathrm{iso-loo-CVC} )
%& := \argmin_{\substack{%
%        \mathbf{a}   \in \mathbb{R}^p \\
%        a_i > a_{i+1} \\
%        \sum_i a_i = \sum_i \hgamma_i^\mathrm{loo-CVC}
%        }}
%\sum_{i=1}^p \left(  a_i - \hgamma_i^\mathrm{loo-CVC}   \right)^2.
%\end{align}
%
\begin{align}
\label{eq:rmt:iso-loocv}
( \hgamma_1^\mathrm{iso-loo-CVC} , \dots, \hgamma_p^\mathrm{iso-loo-CVC} )
& := {\argmin}_{\mathbf{a}   \in \mathbb{R}^p}
{\sum}_{i=1}^p \left(  a_i - \hgamma_i^\mathrm{loo-CVC}   \right)^2,
\end{align}
subject to $a_i > a_{i+1}$ and $ \sum_i a_i = \sum_i \hgamma_i^\mathrm{loo-CVC}$.
This quadratic program is reliably and quickly solved by freely available optimizers such as cvxopt, which was applied in the simulations and experiments below.

%\paragraph{Estimation error results}
To compare CVC and NLS, the simulation  shown in Figure~11 in \citep{ledoit2012nonlinear} is repeated. 20\%, 40\% and 40\% of the eigenvalues are set to 1, 3 and 10, respectively, the data is Gaussian and p/n = 1/3. 
Unfortunately, there is no publicly available code for NLS and we were not able to exactly reproduce the NLS performance  of \cite{ledoit2012nonlinear}. 
Our own implementation, using an open source implementation of Sequential Least Squares Programming \citep{perez2012pyopt,kraft1988software}, is much slower and tends to get stuck in local optima. For precise estimation, it requires multiple initializations and obtained ESEs  are slightly worse for large $p$. 
% Ledoit and Wolf denied to hand out their program code.

\begin{figure} 
\begin{center}
\includegraphics[width= 0.99 \linewidth]{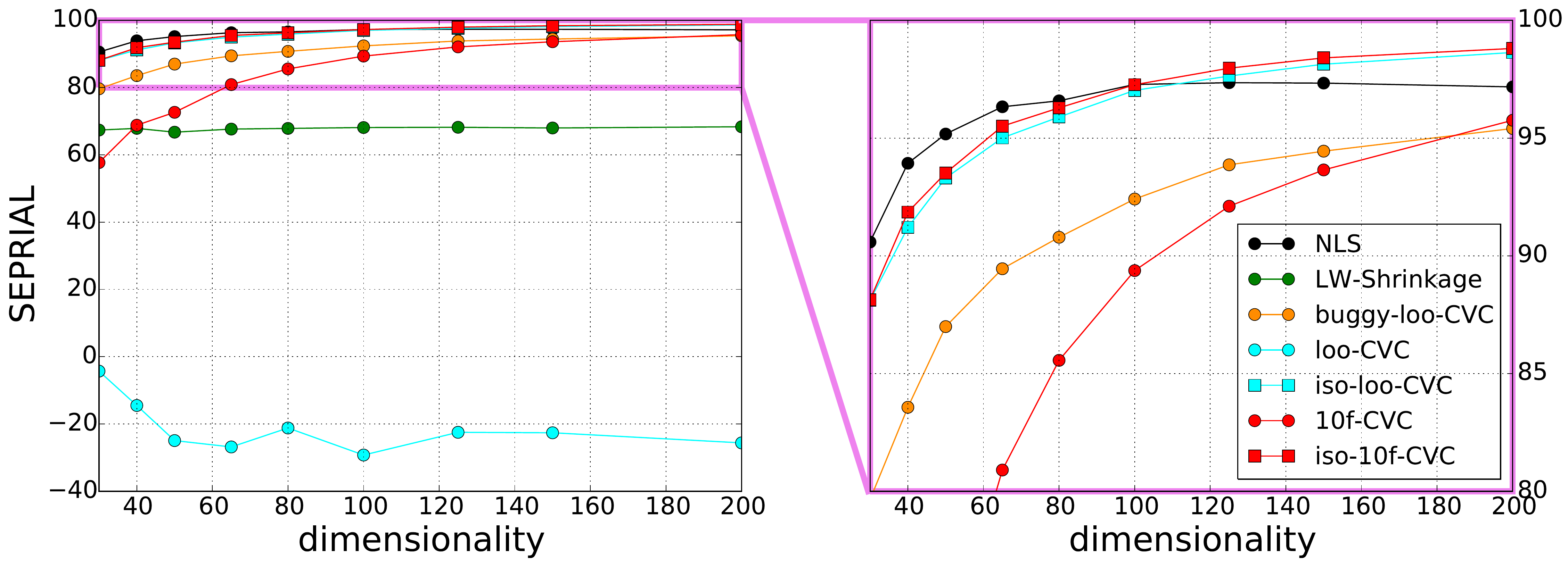}
\caption[Estimation error of CVC estimators.]{Estimation error of CVC estimators. Averaged over 50 repetitions. 
% For stability reasons, 10 repetitions with highest optimization error of NLS were rejected.
}
\label{fig:rmt:performances}
  % generated by sotm_paper_whitening.m
\end{center}
\end{figure}

Figure~\ref{fig:rmt:performances} compares covariance estimators $\hbC$ in terms of 
\emph{sample eigenbasis percentage improvement in average loss} (SEPRIAL):
\begin{align}
\notag
\text{SEPRIAL}\big( \hbC \big) 
:= 100 \cdot \frac{\bbE \| \hbS - \bS^\star \|^2 - \bbE \| \hbC - \bS^\star \|^2 }
{ \bbE \| \hbS - \bS^\star \|^2}.
\end{align} 
Here $\bS^\star$ is the covariance matrix based on the spectrum oracle. The SEPRIAL is 100 for the spectrum oracle and zero for the sample covariance.
The figure shows that loo-CVC yields very bad results, which is in line with the argumentation of the last two paragraphs.
\cite{ledoit2012nonlinear} reported much better results, which were only  reproducable by incorporating a bug into the implementation of loo-CVC: the variance estimates of loo-CVC are calculated from projections ${\hbV^{\mathrm{loo}(t)}_i} {}\transpose \bx_t$  (see eq.\,\eqref{eq:rmt:loocv}). Removing the transpose, ${\hbV^{\mathrm{loo}(t)}_i} {} \bx_t$, defines \emph{buggy-loo-CVC} which exactly yields  the result reported by \cite{ledoit2012nonlinear}. 

At first glance, it is surprising that buggy-loo-CVC is better than the correctly implemented loo-CVC. This is explained by the fact that the simulation of \cite{ledoit2012nonlinear} is performed in the eigenbasis. 
In the eigenbasis, $\bC$ is diagonal, $\hbV$ is approximately the identity and hence $\hbV \transpose \approx \hbV$. 
Using $\hbV$ instead of  $\hbV \transpose$ has the advantage that the variance estimates do not suffer from the  phenomenon illustrated in Figure~\ref{fig:loo_failure}.
Randomly rotating the data before the analysis, buggy-loo-CVC breaks down completely and has much lower SEPRIAL  than loo-CVC.
All estimators besides buggy-loo-CVC are invariant w.r.t.\ the choice of basis.

The cross-validation with 10 folds in \emph{10f-CVC}  yields much higher SEPRIALs than loo-CVC. Both CVC estimators profit from applying isotonic regression: \emph{iso-loo-CVC} and \emph{iso-10f-CVC} achieve competitive SEPRIALs compared to NLS.
For a practical application, the small differences between NLS and iso-10f-CVC should be negligible.

% % % % % % % % % % % % % % % % % % % % % % % % % % % % % % % % % % % % % % % % % % % % % % % % % % % % 
\paragraph{Runtime comparison}
In this section, the runtimes of CVC and NLS are compared. Both approaches are computationally more expensive than LW-Shrinkage:
CVC requires multiple eigendecompositions, which are computationally expensive in high dimensions.
NLS requires an expensive non-linear and non-convex optimization for the estimation of the population eigenvalues.
\cite{ledoit2012nonlinear} state that NLS is superior with respect to runtime, but they do not provide a numerical comparison.

To compare the runtimes of NLS and CVC, the runtimes for both iso-loo-CVC and iso-10f-CVC are measured in the same simulation setting which led to Figure~12 in \cite{ledoit2012nonlinear}. 
As we could not obtain the original program code by Ledoit and Wolf  and our implementation is much slower, a proper comparison of runtimes on the same machine is not possible.
 The NLS runtimes are taken from \citep{ledoit2012nonlinear}, the CVC results are calculated on a 2.3\, GHz Intel i5 Macbook from 2011. The resulting runtimes should be roughly comparable.
\begin{figure}[b]
\begin{center}
\includegraphics[width= 0.6 \linewidth]{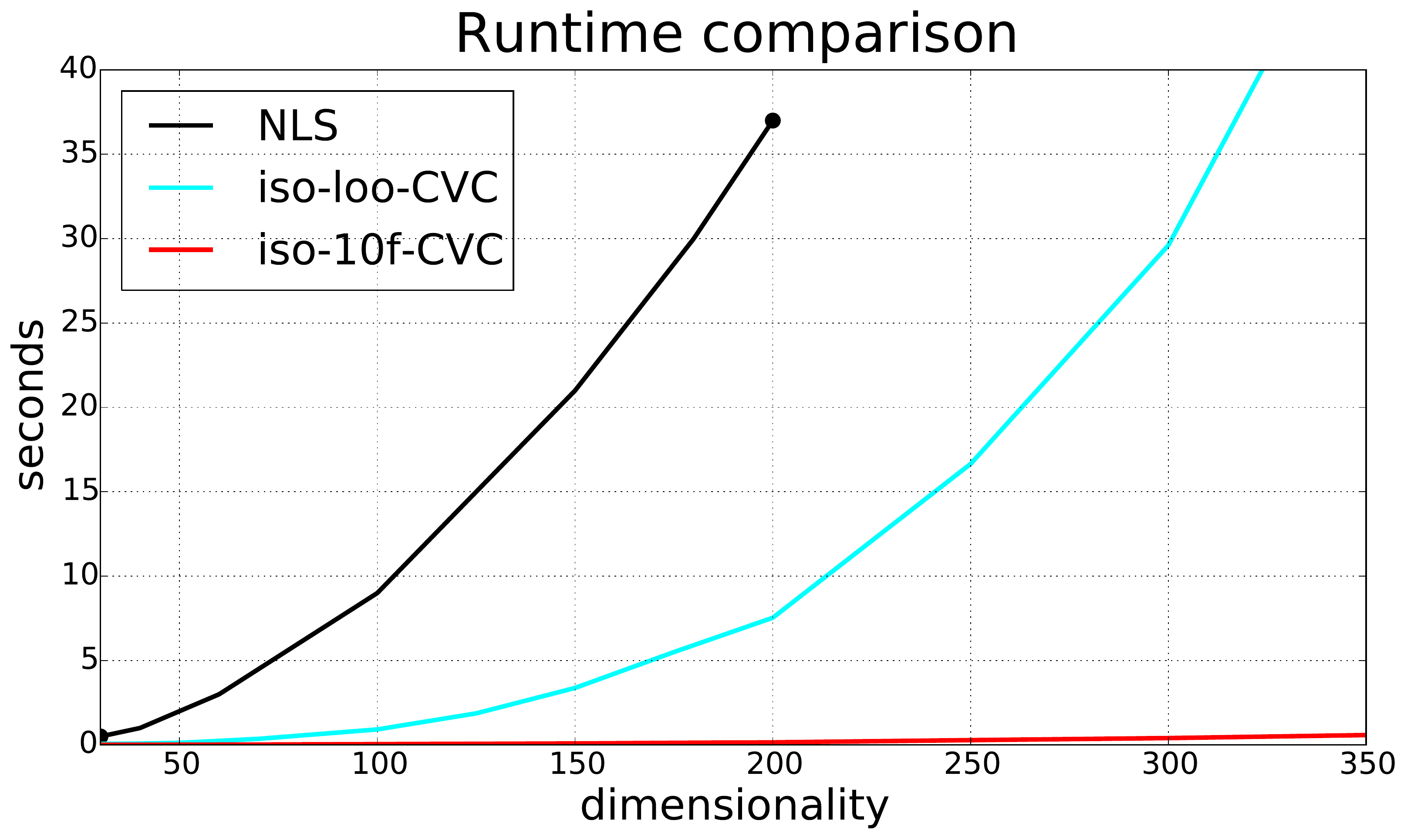}
\caption[Runtime comparison of NLS and CVC.]{Runtime comparison of NLS and CVC.}
\label{fig:runtimes}
  % generated by sotm_paper_whitening.m
\end{center}
\end{figure}
Figure~\ref{fig:runtimes} shows that the runtime for NLS is higher than for the very slow iso-loo-CVC. The cost of {iso-10f-CVC} is negligible compared to the cost of NLS. Contrary to the hypothesis of Ledoit and Wolf, CVC is superior to NLS with respect to runtime.

% The simulations are not entirely fair. As Ledoit and Wolf were not willing to share their code, I could not run the code on the same machine. The NLS results stem from \citep{ledoit2012nonlinear}, the CV results were calculated on 2.3Ghz Intel i5 Macbook from 2011.

% %  % % %  %  % %  % % %  % % %  % % %  %  % %  % % %  % % %  % % %  %  % %  % % %  % % %  % % %  %  % %  % % %  % 
\section{Real world data}
\label{sec:rmt:realworld}
%In this section, the iso-10f-CVC estimator is compared to factor models and shrinkage. aoc-Shrinkage and CV-Shrinkage in the experiments from Section~\ref{sec:reg:rwd}. 
% MI-BCI is not considered, as the CVC estimator is not designed for autocorrelated data.

% %  % % %  %  % %  % % %  % % %  % % %  %  % %  % % %  % % %  % % %  %  % %  % % %  % % %  % % %  %  % %  % % %  % 
%\paragraph{Real world data I: portfolio optimization}
%
\begin{table}
  \caption[CVC results: portfolio risk]{CVC portfolio risk. Mean absolute deviations$\cdot 10^3$ (mean squared
      deviations$\cdot 10^6$) of the resulting portfolios for the
      different covariance estimators and markets. $^\star:=$~significantly better than the other
      models at the 5\% level, tested by a randomization test.}
  \begin{center}
    \begin{tabular}{ l l l l }
      \hline
      \noalign{\vskip \tablerulesep}   
      & US & EU & HK \\
      \hline
      \noalign{\vskip \tablerulesep}   
aoc-Shrinkage & {5.41}$^{}$ ({67.0}$^{}$) & {3.83}$^{}$ ({36.3}$^{}$) & {6.11}$^{}$ ({71.8}$^{}$) \\
DVA-FA & \textbf{5.40}$^{}$ (66.7$^{}$) & 3.84$^{}$ (36.0$^{}$) & 6.12$^{}$ (\textbf{71.7}$^{}$) \\
CVC & 5.40$^{}$ (\textbf{66.7}$^{}$) & \textbf{3.81}$^{\star}$ (\textbf{35.7}$^{\star}$) & \textbf{6.10}$^{}$ (71.8$^{}$) \\
\noalign{\vskip \tablerulesep}   
      \hline
\end{tabular}
  \end{center}
  \label{tab:spec:pfolio}
\end{table}
We repeat a \emph{portfolio optimization}   experiment from \cite{Bar13} and compare to DVA Factor Analysis, the covariance estimator with the best performance. In addition, we include aoc-Shrinkage, the best performing Shrinkage approach from \cite{BarMue13}. Table~\ref{tab:spec:pfolio} compares the results.
Although the competing approaches explicitly model the factor structure in the data,
the CVC estimator is on par for the US and HK market and even significantly reduces portfolio risk for the EU market.

As a second real world data experiment, we repeat a Neuro-Usability experiment from \cite{BarMue13}, where Linear Discriminant Analysis based on different covariance estimators is used to classify EEG recordings of noisy and noise-free phonemes. Artificial noise has been injected into one of the electrodes. 
In the experiment, aoc-Shrinkage outperformed other Shrinkage approaches. We include covariance matrices based on Probabilistic PCA, Factor Analysis and DVA Factor Analysis into the comparison. 
Table~\ref{tab:spec:bci} shows that CVC significantly outperforms aoc-Shrinkage and the factor models. As aoc-Shrinkage, CVC is not affected by the injected noise. 
For the EEG data set, the factor models do not yield competitive performance: the data set does not have a pronounced factor structure.

We do not compare CVC to NLS on real world data because (i) the runtime of NLS is not feasible and (ii) the numerical stability of the optimization is not sufficient.
\begin{table}
 \caption[CVC, Shrinkage and FM accuracies for the Neuro-Usability ERP-BCI data]{CVC, Shrinkage and FM accuracies for classification tasks on the Neuro-Usability ERP-BCI data. Artificially injected noise in one electrode. $^\star:=$ significantly better than all other methods at the 5\% level, tested by a randomization test.}
\centering
\begin{tabular}{ l l l l l l l }
\hline
\noalign{\vskip \tablerulesep}   
%  & \multicolumn{6}{l}{anneN}  \\
$\sigma_{noise}$ & 0 & 10 & 30 & 100 & 300 & 1000 \\
\hline
\noalign{\vskip \tablerulesep}   
% LDA & & & &  & & \\
% \quad accuracy  & 92.24\%  & 92.23\%  & 92.23\%  & 92.24\%  & 92.24\%  & 92.24\% \\
CVC-LDA   & $\mathbf{93.74\%}^\star$  & $\mathbf{93.73\%}^\star$  & $\mathbf{93.73\%}^\star$  & $\mathbf{93.73\% }^\star$  & $\mathbf{93.73\% }^\star$  & $\mathbf{93.73\% }^\star$ \\
\noalign{\vskip \tablerulesep}   
aoc-Shrinkage-LDA  & ${93.27\% }$  & ${93.27\% }$  & ${93.24\% }$  & ${92.88\% }$  & ${93.16\% }$  & ${93.19\% }$ \\
PPCA-LDA  & 91.13\%  & 91.15\%  & 91.11\%  & 91.13\%  & 91.07\%  & 90.85\% \\
FA-LDA   & 91.61\%  & 91.6\%  & 91.46\%  & 91.14\%  & 90.4\%  & 90.35\% \\
DVA-FA-LDA   & 91.96\%  & 91.93\%  & 91.17\%  & 89.44\%  & 90.05\%  & 89.71\% \\
\hline
\end{tabular}
\label{tab:spec:bci}
\end{table}

% % % % % % % % % % % % % % % % % % % % % % % % % % % % % % % % % % % % % % % % % % % % % % % % % % % % % % % % % % % % % % % 
\section{Summary}
\label{sec:rmt:summary}
Spectrum correction methods keep the sample eigenvectors and only modify the eigenvalues. The state-of-the-art is Nonlinear Shrinkage (NLS), 
a highly complex method from Random Matrix Theory which minimizes the expected squared error.
This article proposed a cross-validation-based covariance estimator (CVC)  based on isotonic regression and 10-fold cross-validation which optimizes the same loss function.
It  yields competitive results at greatly reduced complexity and runtime.

Both NLS and CVC are computationally more complex than LW-Shrinkage. The non-linear and non-convex optimization required by NLS is very time-consuming. 
A comparison in the simulation setting considered by \cite{ledoit2012nonlinear} showed that the cost of iso-loo-CVC is significantly smaller,  although it requires $n+1$ eigendecompositions.
Using ten cross-validation folds greatly reduces the computational cost  and renders CVC applicable in practice. Simulations show that the estimation errors of CVC and NLS are on the same level. 

%\paragraph{Performance}
On all data sets considered, the CVC estimator yields better or equal performance than state-of-the-art covariance estimators based on factor models or shrinkage. This shows that CVC is an excellent general purpose covariance estimator. The results are especially impressive as the competing methods are specifically tailored to the applications.
%. We are convinced that 

%\paragraph{Runtime}

%\subsubsection*{Acknowledgments}
%
%Use unnumbered third level headings for the acknowledgments. All
%acknowledgments go at the end of the paper. Do not include
%acknowledgments in the anonymized submission, only in the final paper.

\newpage

\small
\bibliography{ida,machineLearning,finance,dbML}{}

\newcommand{\etalchar}[1]{$^{#1}$}
 \newcommand{\noop}[1]{}
\begin{thebibliography}{BHH{\etalchar{+}}13}

\bibitem[BBBB72]{barlow1972statistical}
Richard~E Barlow, David~J Bartholomew, JM~Bremner, and H~Daniel Brunk.
\newblock {\em Statistical inference under order restrictions: the theory and
  application of isotonic regression}.
\newblock Wiley New York, 1972.

\bibitem[BHH{\etalchar{+}}13]{Bar13}
Daniel Bartz, Kerr Hatrick, Christian~W. Hesse, Klaus-Robert {M\"uller}, and
  Steven Lemm.
\newblock Directional {Variance} {Adjustment}: Bias reduction in covariance
  matrices based on factor analysis with an application to portfolio
  optimization.
\newblock {\em PLoS ONE}, 8(7):e67503, 07 2013.

\bibitem[BM13]{BarMue13}
Daniel Bartz and Klaus-Robert M\"{u}ller.
\newblock Generalizing analytic shrinkage for arbitrary covariance structures.
\newblock In C.J.C. Burges, L.~Bottou, M.~Welling, Z.~Ghahramani, and K.Q.
  Weinberger, editors, {\em Advances in Neural Information Processing Systems
  26}, pages 1869--1877. Curran Associates, Inc., 2013.

\bibitem[DGL13]{devroye2013probabilistic}
Luc Devroye, L{\'a}szl{\'o} Gy{\"o}rfi, and G{\'a}bor Lugosi.
\newblock {\em A probabilistic theory of pattern recognition}, volume~31.
\newblock Springer Science \& Business Media, 2013.

\bibitem[DS85]{dey1985estimation}
Dipak~K Dey and C~Srinivasan.
\newblock Estimation of a covariance matrix under {S}tein's loss.
\newblock {\em The Annals of Statistics}, pages 1581--1591, 1985.

\bibitem[EK08]{Kar08}
Noureddine El~Karoui.
\newblock Spectrum estimation for large dimensional covariance matrices using
  random matrix theory.
\newblock {\em Annals of Statistics}, 36(6):2757--2790, 2008.

\bibitem[ER05]{EdeRao05}
Alan Edelman and N.~Raj Rao.
\newblock Random matrix theory.
\newblock {\em Acta Numerica}, 14:233--297, 2005.

\bibitem[K{\etalchar{+}}88]{kraft1988software}
Dieter Kraft et~al.
\newblock {\em A software package for sequential quadratic programming}.
\newblock DFVLR Obersfaffeuhofen, Germany, 1988.

\bibitem[KR99]{kearns1999algorithmic}
Michael Kearns and Dana Ron.
\newblock Algorithmic stability and sanity-check bounds for leave-one-out
  cross-validation.
\newblock {\em Neural Computation}, 11(6):1427--1453, 1999.

\bibitem[LCPB00]{Lal00}
Laurent Laloux, Pierre Cizeau, Marc Potters, and Jean-Phillipe Bouchaud.
\newblock Random matrix theory and financial correlations.
\newblock {\em International Journal of Theoretical and Applied Finance},
  3(3):391--397, 2000.

\bibitem[LP11]{ledoit2011eigenvectors}
Olivier Ledoit and Sandrine P{\'e}ch{\'e}.
\newblock Eigenvectors of some large sample covariance matrix ensembles.
\newblock {\em Probability Theory and Related Fields}, 151(1-2):233--264, 2011.

\bibitem[LW03]{LedWol03}
Olivier Ledoit and Michael Wolf.
\newblock Improved estimation of the covariance matrix of stock returns with an
  application to portfolio selection.
\newblock {\em Journal of Empirical Finance}, 10:603--621, 2003.

\bibitem[LW04]{LedWol04}
Olivier Ledoit and Michael Wolf.
\newblock A well-conditioned estimator for large-dimensional covariance
  matrices.
\newblock {\em Journal of Multivariate Analysis}, 88(2):365--411, 2004.

\bibitem[LW{\etalchar{+}}12]{ledoit2012nonlinear}
Olivier Ledoit, Michael Wolf, et~al.
\newblock Nonlinear shrinkage estimation of large-dimensional covariance
  matrices.
\newblock {\em The Annals of Statistics}, 40(2):1024--1060, 2012.

\bibitem[LW14]{ledoit2014spectrum}
Olivier Ledoit and Michael Wolf.
\newblock Spectrum estimation: A unified framework for covariance matrix
  estimation and pca in large dimensions.
\newblock {\em arXiv preprint arXiv:1406.6085}, 2014.

\bibitem[MP67]{MarPas67}
Vladimir~A. Mar\v{c}enko and Leonid~A. Pastur.
\newblock Distribution of eigenvalues for some sets of random matrices.
\newblock {\em Mathematics of the USSR-Sbornik}, 1(4):457, 1967.

\bibitem[PJM12]{perez2012pyopt}
Ruben~E Perez, Peter~W Jansen, and Joaquim~RRA Martins.
\newblock pyopt: a python-based object-oriented framework for nonlinear
  constrained optimization.
\newblock {\em Structural and Multidisciplinary Optimization}, 45(1):101--118,
  2012.

\bibitem[RPGS02]{Ros02}
B.~Rosenow, V.~Plerou, P.~Gopikrishnan, and H.~E. Stanley.
\newblock Portfolio optimization and the random magnet problem.
\newblock {\em Europhysics Letters}, 59(4):500--506, 2002.

\bibitem[Sil95]{silverstein1995strong}
Jack~W Silverstein.
\newblock Strong convergence of the empirical distribution of eigenvalues of
  large dimensional random matrices.
\newblock {\em Journal of Multivariate Analysis}, 55(2):331--339, 1995.

\bibitem[Ste56]{Ste56}
Charles Stein.
\newblock Inadmissibility of the usual estimator for the mean of a multivariate
  normal distribution.
\newblock In {\em Proc. 3rd Berkeley Sympos. Math. Statist. Probability},
  volume~1, pages 197--206, 1956.

\bibitem[Ste86]{stein1986lectures}
Charles Stein.
\newblock Lectures on the theory of estimation of many parameters.
\newblock {\em Journal of Soviet Mathematics}, 34(1):1373--1403, 1986.

\bibitem[VK82]{vapnik1982estimation}
Vladimir~Naumovich Vapnik and Samuel Kotz.
\newblock {\em Estimation of dependences based on empirical data}, volume~40.
\newblock Springer-verlag New York, 1982.

\end{thebibliography}


 \newcommand{\noop}[1]{}
\begin{thebibliography}{BBM08}

\bibitem[BBM08]{braun2008relevant}
Mikio~L Braun, Joachim~M Buhmann, and Klaus-Robert M{\"u}ller.
\newblock On relevant dimensions in kernel feature spaces.
\newblock {\em The Journal of Machine Learning Research}, 9:1875--1908, 2008.

\end{thebibliography}
\end{document}


\maketitle

% % % % % % % % % % % % % % % % % % abstract
% % % % % % % % % % % % % % % % % % abstract
% % % % % % % % % % % % % % % % % % abstract
%\begin{abstract} 
%Many machine learning algorithms require precise estimates of covariance matrices. The sample covariance matrix performs poorly in high-dimensional settings, which has stimulated the development of alternative methods, the majority based on factor models and shrinkage. Recent work of Ledoit and Wolf has extended the shrinkage framework to Nonlinear Shrinkage (NLS), a more powerful covariance estimator based on Random Matrix Theory.
%
%Our contribution shows that, contrary to claims in the literature, cross-validation based covariance matrix  estimation (CVC) yields comparable performance at strongly reduced complexity and runtime. On two real world data sets, we show that the CVC estimator yields superior results than competing shrinkage and factor based methods.
%\end{abstract}

%
% % % % % % % % % % % % % % % % % % % % % % % % % % % % % % % % % % % % % % % % % % % % % % % % % % % % % % % % % % % % % % % 

\section{Discriminative information and covariance}
\label{sec:sys:intro}
\begin{figure}
\begin{center}
\includegraphics[width= 0.99 \linewidth]{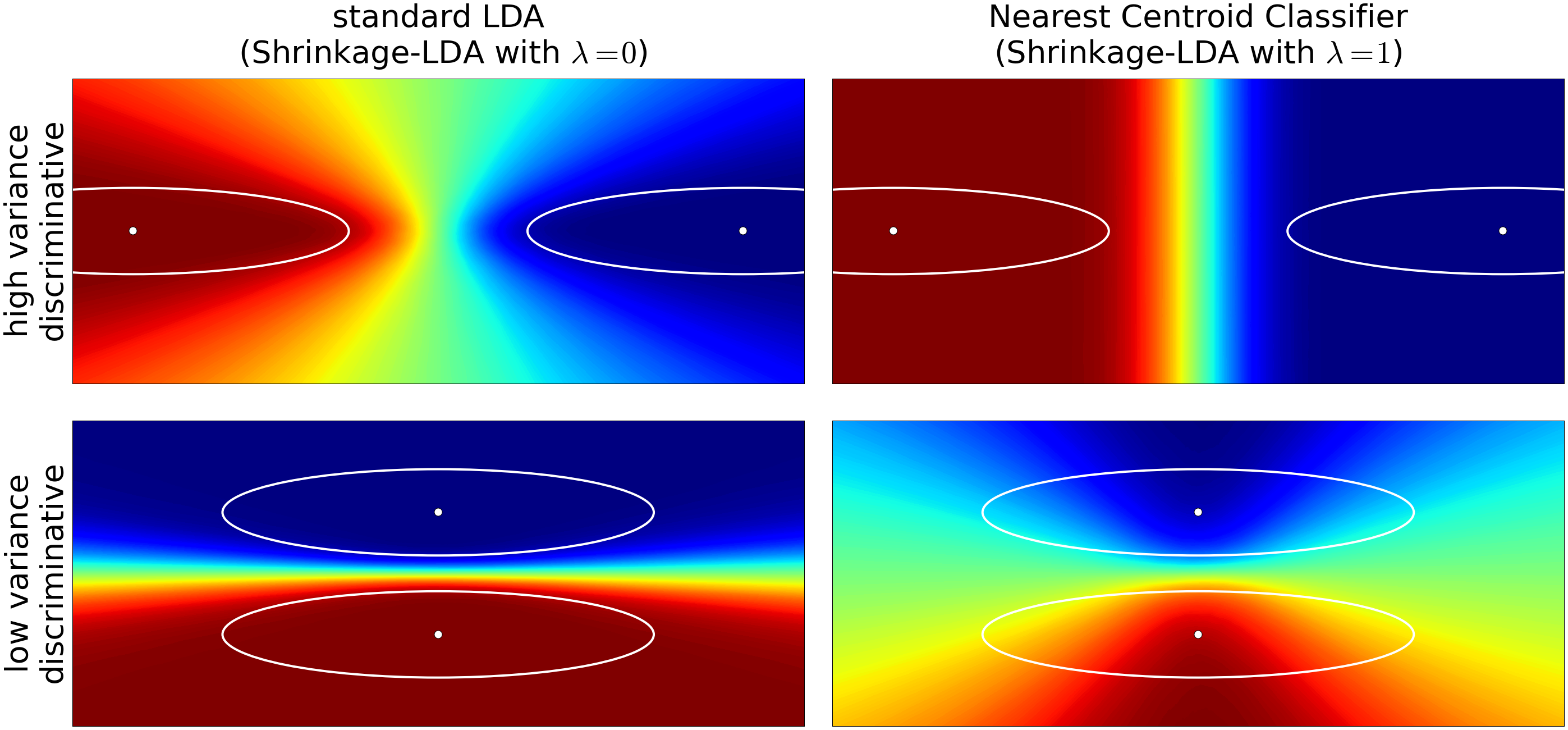}
\caption[2D illustration: average predictions]{2D illustration: average predictions. The color coding indicates the probability that an LDA classifier based on five observations per class classifies a point as red or blue. Dark blue and dark red indicate 100\% probability of being classified as class blue and class red, respectively.}
\label{fig:2d_lda}
\end{center}
\end{figure}

For Linear Discriminant Analysis (LDA), the orientation of the covariance relative to the location of the sample means is decisive.  
Figure~\ref{fig:2d_lda}  illustrates this in a two-dimensional setting with five observations for each of the two classes red and blue. The upper and lower row show two different settings: for the upper row, the mean differences lie in the direction of high variance, as illustrated by the constant likelihood contours of the distributions of the two classes. For the lower row, the mean differences lie in the direction of low variance. 

The two columns show two different classification approaches. The left column shows \emph{standard LDA}, equivalent to Shrinkage-LDA with $\lambda=0$, while the right shows the \emph{Nearest Centroid Classifier}, equivalent to Shrinkage-LDA with $\lambda=1$.

The color coding shows classification probabilities for the two classes. Interestingly, the picture is very different for the two settings: if the high-variance direction is discriminative,  the Nearest Centroid Classifier is much better, if the low-variance direction is discriminative, LDA gives better decision boundaries.

\begin{figure}
\begin{center}
\includegraphics[width= 0.99 \linewidth]{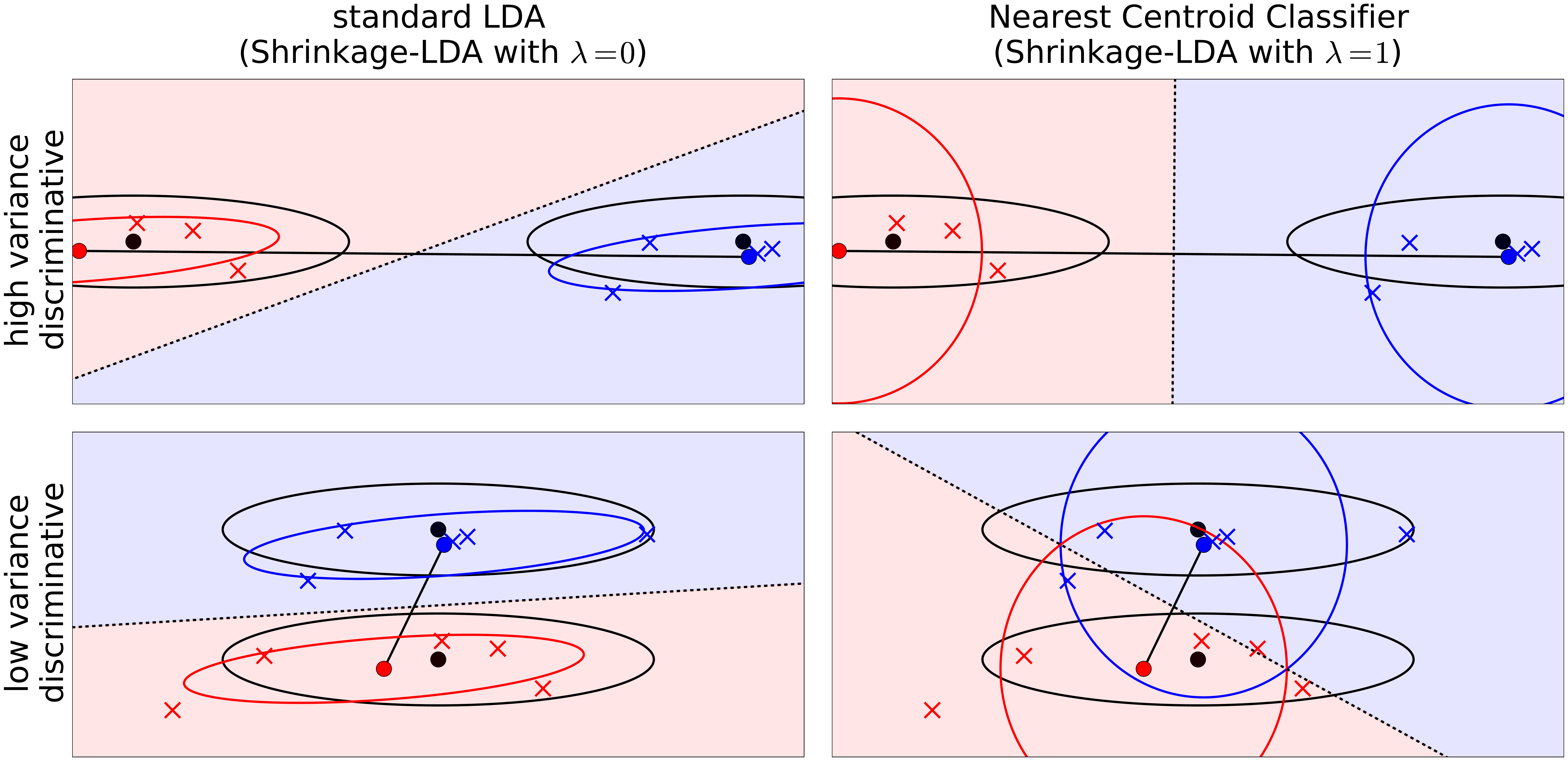}
\caption{2D illustration: single realization.}
\label{fig:2d_lda_errors}
\end{center}
\end{figure}

To understand this, consider Figure~\ref{fig:2d_lda_errors}. If the direction of high variance is discriminative, relatively small errors in shape of the covariance estimate can lead to highly suboptimal decision boundaries. If the direction of low variance is discriminative, errors in the mean in the direction of high variance adversely affect nearest centroids.

The illustrations deliver an important message: 
from the point of view of covariance estimation, the two settings are indistinguishable.
Although very different covariance estimators are optimal, the covariance estimates of  methods like LW-Shrinkage, NLS and CVC will not differ. 
This is especially dramatic for LW-Shrinkage-LDA which yields an interpolation between standard LDA and nearest centroids  and cannot be optimal in both settings.

\paragraph{Discriminative information and covariance in more than two dimensions}
In the simulation above, either the low- or the high-variance direction was discriminative. In the simulations below, the following generalization to higher dimensions is considered:
\begin{itemize}
\item
\emph{high-variance directions are discriminative:}
it is often assumed that the directions of high variance are informative. This does not mean that the Nearest Centroid Classifier is optimal, because the discriminative high-variance directions may vary in strength.
An alternative definition of this setting is that the discriminative directions are those which dominate the Frobenius norm of the covariance matrix.
\item
\emph{low-variance directions are discriminative:} 
low-variance directions are all directions which are dominated by the strongest eigendirections.
Note that this definition does not only include the directions of lowest variance, which are often discarded in practice as they do not carry discriminative information. 
An alternative definition of this setting is that the directions which dominate the Frobenius norm of the covariance matrix are not discriminative.
%
%\item
%\emph{spread-out discriminative information:}
%in this setting, the discriminative information is not concentrated in the subspace of high or low variance. 
\end{itemize}
%
%\cite{braun2008relevant} studied the relationship between discriminative information and directions of high variance for the case of kernel principal component analysis. They proved that the discriminative information is contained in the leading components if the kernel matches the learning problem.  Note that this theoretical result does not pose a restriction on the location of the discriminative information in the LDA setting: the linear kernel, tightly related to the covariance matrix, only matches the learning problem if the discriminative information is contained in the  leading components. If the discriminative information is not restricted to the leading components, the theoretical result is not applicable without further assumptions.

% % % % % % % % % % % % % % % % % % % % % % % % % % % % % % % % % % % % % % % % % % % % % % % % % % % % % % % % % % % % % % % 

\section{Simulation design}
\label{sec:sys:simdesign}
The last section has illustrated that the orientation of the covariance relative to the discriminative directions plays an important role in Linear Discriminant Analysis. This section contains a systematic analysis.

\begin{figure}
\begin{center}
\includegraphics[width= 0.75 \linewidth]{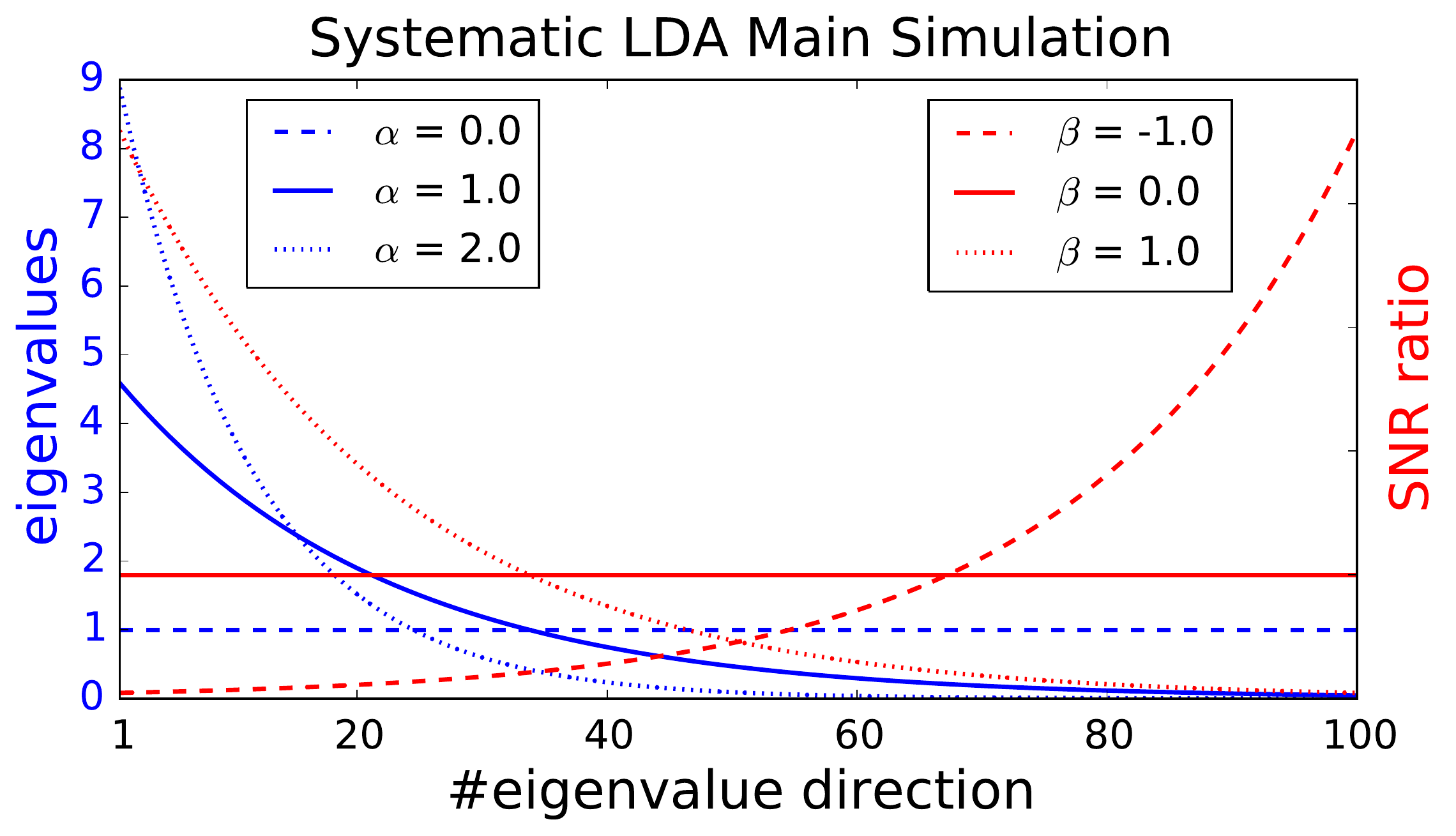}
%\caption{SNR and eigenvalues in the main simulation and variation 1.}
\caption{SNR and eigenvalues in the simulation.}
\label{fig:illu_sim1}
\end{center}
\end{figure}

For the simulation, Gaussian data with two classes is generated ($p=100,n=200$). The eigenvalues $\gamma_i$ are chosen logarithmically spaced between $10^{-\alpha}$ and $10^{\alpha}$ and values of $\alpha$ between 0 and 2 are considered. For $\alpha=0$ the spectrum is flat, for $\alpha=2$ the spectrum is heavily tilted (see Figure~\ref{fig:illu_sim1}).
%\begin{align}
%\label{eq:sys:sim1ev}
%\gamma_i := c_{\alpha} 10^{\alpha (p-2(i-1))/p  }
%\end{align}

The mean of class A is set to $\bmu^{A} = 0$, while the entries in $\bmu^{B}$ are defined in the following way: 
$ \mu_i^{B} / c_\mathrm{90\%} / \sqrt{\gamma_i} $ are logarithmically spaced between $10^{\beta}$ and $10^{-\beta}$,
where $c_\mathrm{90\%}$ is chosen such that the Bayes optimal classifier obtains 90\% accuracy.

For $\beta$, values between -1 and 1 are considered. For $\beta < 0$ the Signal-to-Noise ratio is high in the directions of low variance, for $\beta = 0$ the SNR is identical in all directions and for $\beta > 0$ the SNR is high in the directions of low variance (see Figure~\ref{fig:illu_sim1}).

%Figure~\ref{fig:illu_sim1} shows the Eigenvalues and the SNR. For $\alpha = 0$, the eigenvalues are all equal to one, for the $\alpha =2$ the eigenvalues are spread between 0.01 and 100. For negative $\beta$, the SNR is high in the low variance regions while for positive $\beta$ the SNR is high in the high variance regions. For $\beta=0$, high and low variance regions are equally discriminitive.

%\paragraph{Variation 1: low Bayes-optimal classification rate}
%As a variation of the main simulation, the Bayes-optimal classification rate of  was set to  60\%. This increases the noise in the cross-validation of the optimal shrinkage intensity. Apart from that, the simulation was identical to the previous one.

%\paragraph{Variation 2: non-monotonous SNR}
%The second variation considered a slightly artificial case: the information is concentrated either in the medium-variance range or  only in the high and low-variance directions. 
%Variation 2 only differs in the definition of $\bmu^{B}$. The first and the second half of the entries in $\bmu^{B}$ are defined such that
%the corresponding $\mu_i^{B} / c_\mathrm{90\%} / \sqrt{\gamma_i} $
%are logarithmically spaced between $10^{\beta}$ and $10^{-\beta}$ 
%and
%$10^{-\beta}$ and $10^{\beta}$, respectively.
%
%\begin{align}
%\label{eq:sys:sim2snr}
%\mu_i^{B} := c_\mathrm{90\%} 10^{ - \beta_i  2|i-p/2|/p} \sqrt{\gamma_i},
%\end{align}
%Figure~\ref{fig:illu_sim2} shows the distribution of the eigenvalues and the SNR.
%%
%\begin{figure}
%\begin{center}
%\includegraphics[width= 0.75 \linewidth]{images_supplemental/illustration_sim2.pdf}
%\caption{SNR and eigenvalues in variation 2.}
%\label{fig:illu_sim2}
%\end{center}
%\end{figure}

% % % % % % % % % % % % % % % % % % % % % % % % % % % % % % % % % % % % % % % % % % % % % % % % % % % % % % % % % % % % % % % 

\section{Simulation results}
\label{sec:sys:simresults}
\begin{figure}
\begin{center}
\includegraphics[width= 0.99 \linewidth]{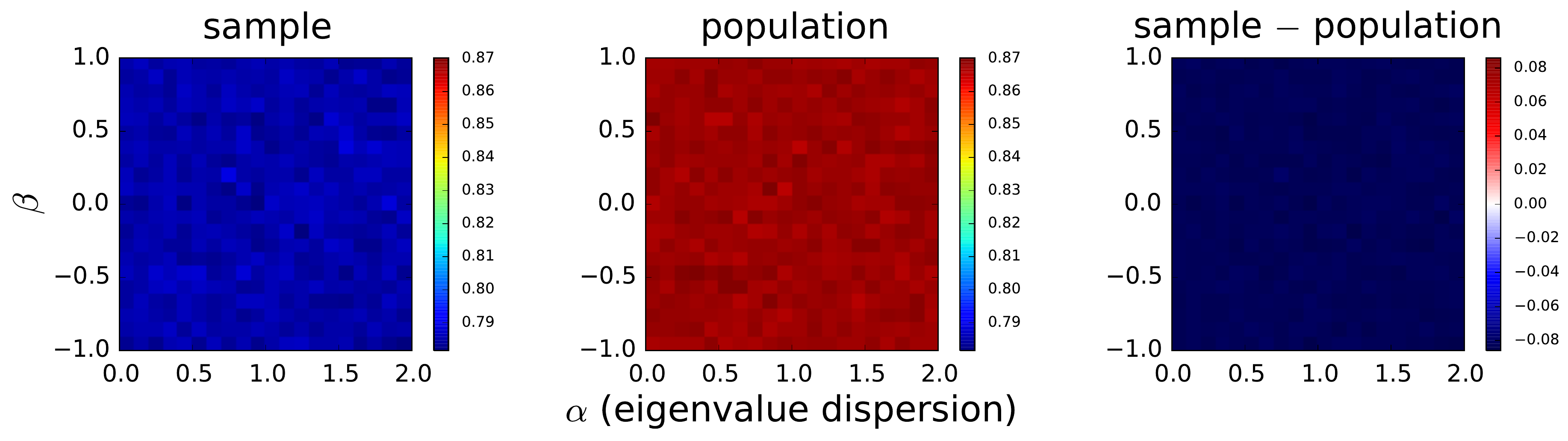} \\
\includegraphics[width= 0.99 \linewidth]{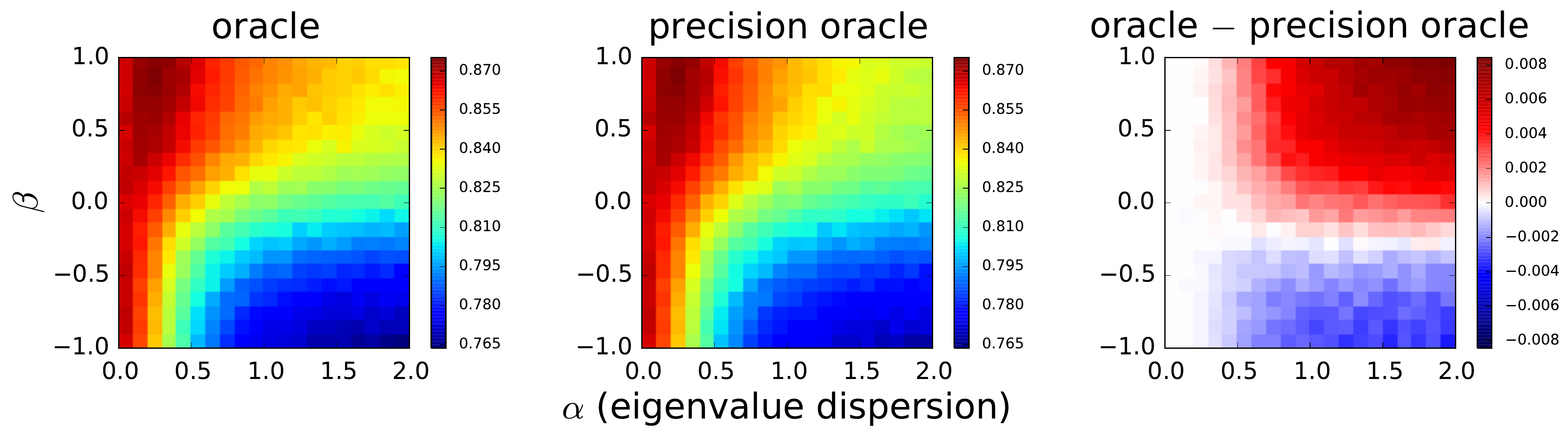}\\
\includegraphics[width= 0.99 \linewidth]{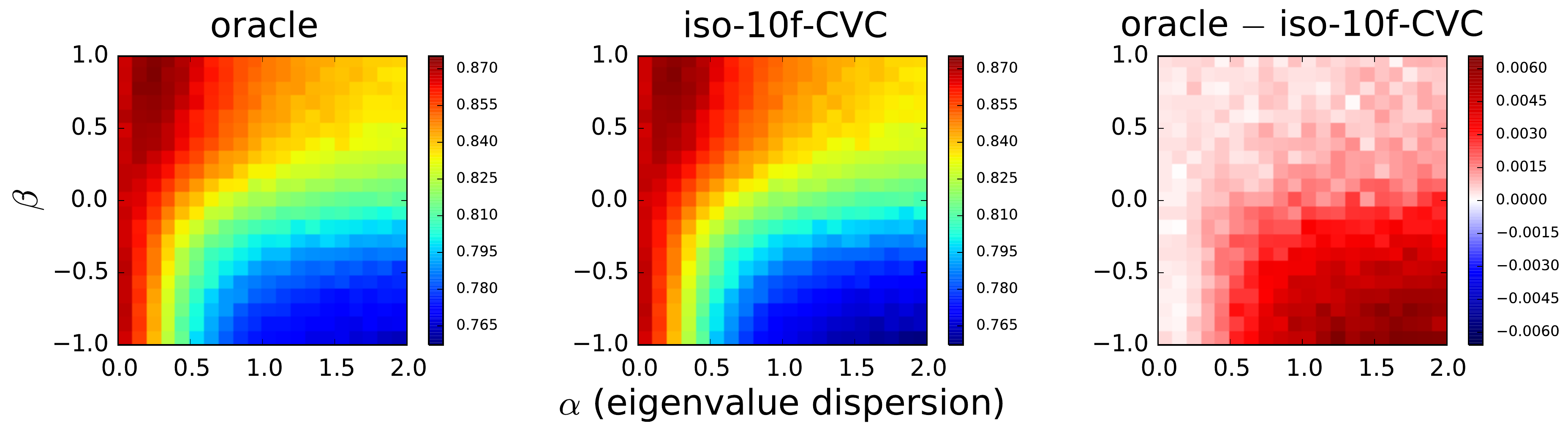} \\
\includegraphics[width= 0.99 \linewidth]{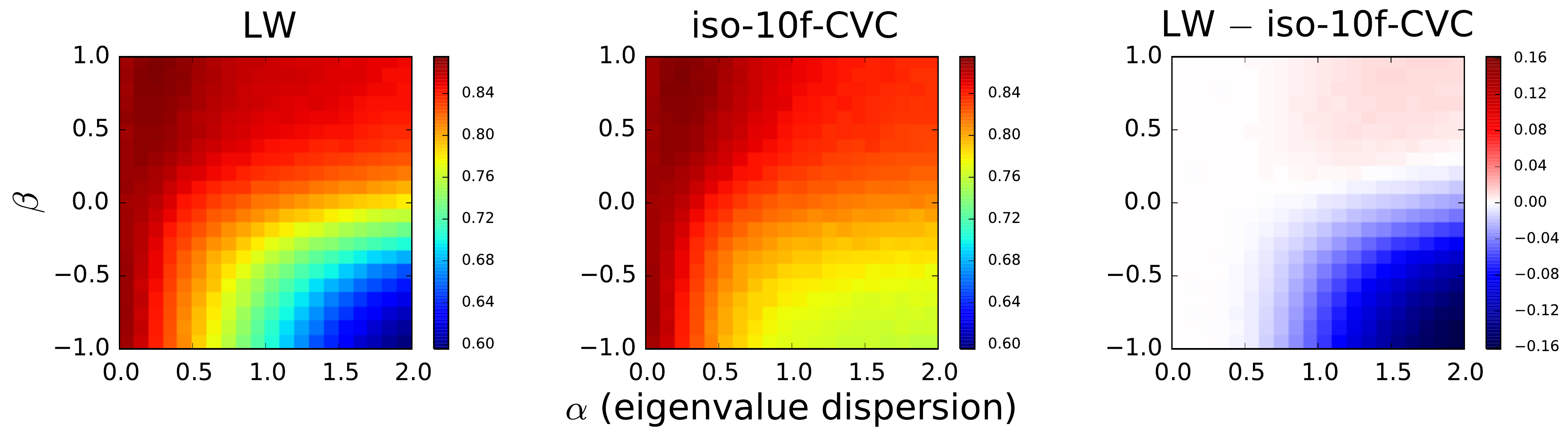} \\
\caption{Systematic comparison of classification accuracies.}
\label{fig:sys:all}
\end{center}
\end{figure}
Figure~\ref{fig:sys:all} shows the results of the main simulation. The systematic analysis yields answers to  a set of questions.

\paragraph{How good are sample and population covariance?}
Figure~\ref{fig:sys:all} (first row) shows that the performance is independent of the location and shape parameters $\alpha$ and $\beta$.
The population covariance achieves an accuracy of approximately 87\%. Compared to the Bayes optimal classifier, 3\% are lost due to the estimation error in the sample means. The sample covariance is about 8\% worse than the population covariance.

\paragraph{Is the precision oracle better than the (spectrum) oracle?}
In the article, two optimal estimators in the sample eigenbasis were discussed: the (spectrum) oracle eq.\,(6) has minimum squared error with respect to covariance, the precision oracle eq.\,(11) has minimum squared error with respect to the precision matrix. 

Figure~\ref{fig:sys:all} (second row) shows that the results are very similar. Classification accuracies differ by less than 1\%.  If high-variance directions are discriminative or the information is spread evenly, the spectrum oracle is superior. If directions of low variance are discriminative, the precision oracle is superior.

To understand this behavior, consider the oracles in the population eigenbasis. Then, we have from eq.\,(6) and eq.\,(11):
\begin{align}
\label{eq:rmt:optimal_spectrum_eigen}
\gamma^\star_i 
= \sum_j \hv_{ji}^2 \gamma_j
 \qquad  
\mathrm{and}
 \qquad  
\gamma^\diamond_i = \frac{1}
{
 \sum_{j} \hv^2_{ji} \gamma_i^{-1}
}.
\end{align}

\begin{figure}
\begin{center}
\includegraphics[width= 0.8 \linewidth]{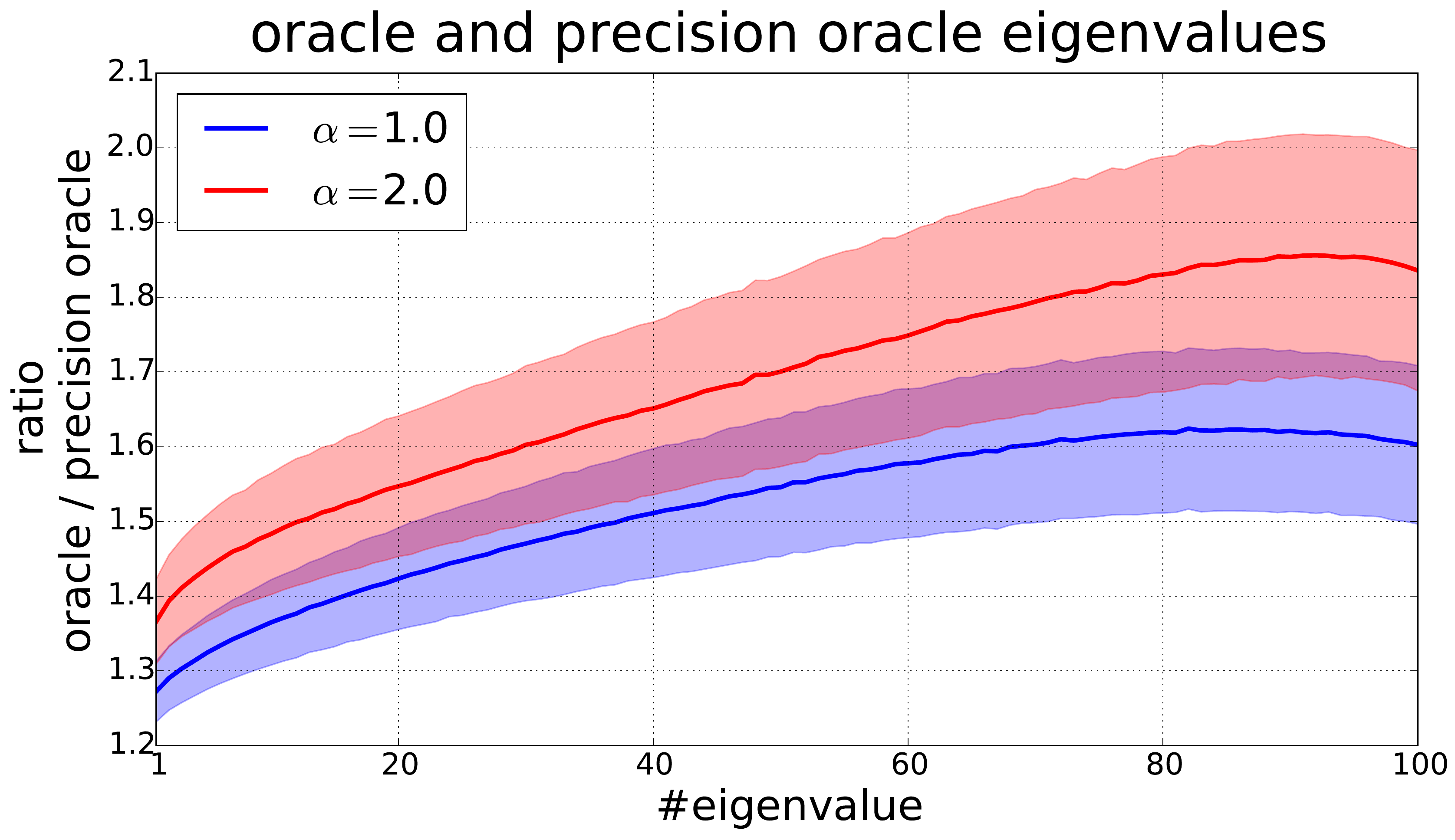}
\caption{Comparison of (spectrum) oracle and precision oracle.}
\label{fig:sys:spec_oracles}
\end{center}
\end{figure}

The oracle is a weighted arithmetic mean of the eigenvalues, while the precision oracle is a weighted harmonic mean. The harmonic mean is always smaller than the arithmetic mean, but the \emph{ratio} differs for large and small eigenvalues.
Figure~\ref{fig:sys:spec_oracles} shows that the precision oracle yields, relative to the oracle, very small eigenvalues for the small sample eigenvalues. Hence it favors the directions of small variance and is superior when these directions are discriminative.

\paragraph{Does the CVC estimator achieve performance comparable to the oracle?}
Figure~\ref{fig:sys:all} (third row) shows that the differences in accuracy are smaller than 1\%. If directions of high variance are discriminative, the difference is close to zero, it is largest for discriminative low-variance directions.

\paragraph{How does LW-Shrinkage compare to CVC?}
Figure~\ref{fig:sys:all} (last row) shows that both estimators are on par when the spectrum is flat. For a strongly tilted spectrum, LW-Shrinkage performs slightly better when the directions of high variance are discriminative. When the directions of low variance are discriminative, CVC performs much better, in the extreme case the difference in accuracies is at 20\%.

%\begin{figure}
%\begin{center}
%\includegraphics[width= 0.95 \linewidth]{images_supplemental/sys_performance__n_200_p_100_R_100_accB_60_linear_False_mono_True_blob_False_so_0_CV_iso-10f-CVC.pdf}
%\caption{Variation 1: systematic comparison  for low Bayes-optimal accuracy.}
%\label{fig:sys:lowbayes}
%\end{center}
%\end{figure}

%\begin{figure}
%\begin{center}
%\includegraphics[width= 0.95 \linewidth]{images_supplemental/sys_performance__n_200_p_100_R_100_linear_False_mono_False_so_0_CV_iso-10f-CVC.pdf}
%\caption{Variation 2: systematic comparison for non-monotonous SNR.}
%\label{fig:sys:nmsnr}
%\end{center}
%\end{figure}

\paragraph{Limitations of the simulations} 
Although the  LDA setting is relatively simple, the analysis presented in this section cannot be complete: it focused on the eigenvalue dispersion and the orientation of the covariance and had to neglect other aspects:
\begin{itemize}
\item 
\emph{Concentration $c$, dimensionality $p$ and number of observations $n$:}
in the above analysis, $p = 100$ and $n = 200$ were fixed. At lower concentration $c=n/p$ higher regularization is optimal. At smaller $n$ all estimators get less precise.
\item 
\emph{Shape of the spectrum:}
the logarithmic spacing of the population eigenvalues, displayed in Figure~\ref{fig:illu_sim1}, is just one possible parametrization. Besides other functional forms there could also be a factor structure with single eigenvalues clearly separated from the bulk.
\item 
\emph{Shape of the SNR:}
% The functional form in eq.\,\eqref{eq:sys:sim1snr} and~\eqref{eq:sys:sim2snr}, 
the locations of the means which were chosen for the simulations yield the SNRs displayed in Figure~\ref{fig:illu_sim1}. 
Other locations  of the means leading to differently shaped SNR curves are possible.
\item 
\emph{Effect of the distribution:}
real world data is often non-Gaussian. It would be interesting to analyze the effect a distribution with heavy tails has on the results.
\item 
\emph{Number of classes:}
the systematic analysis only considered two classes. An extension to multiple classes would greatly increase the degrees of freedom in the analysis.
\end{itemize}

%\paragraph{Relation to experiments on real world data}
%In Sections~\ref{sec:reg:rwd} and~\ref{sec:rmt:realworld}, Linear Discriminant Analysis has been applied to  the MSM-ERP-BCI, ISOLET and USPS data sets. The systematic analysis gives us insights into the results reported there. 
%
%In general, the discriminative information in these experiments seems to be restricted to the high-variance dimensions: LW-Shrinkage yields good results, although CVC tends to be a bit better.
%In addition, the Bayes-optimal classification rate is relatively high in all experiments. 
%These two facts explain why the best results are obtained by CV-Shrinkage.
%
%In none of the experiments the discriminative  information is spread evenly over large regions of the spectrum, the setting in which CVC would outperform CV-Shrinkage. This does not rule out the existence of such  settings in other application domains.

% % % % % % % % % % % % % % % % % % % % % % % % % % % % % % % % % % % % % % % % % % % % % % % % % % % % % % % % % % % % % % % 
\section{Supplementary Material summary}
\label{sec:sys:summary}
The performance in a practical application is the main uncertainty in covariance estimation research. Most covariance estimators     optimize a measure of estimation error, which may or may not be a good proxy for the application-dependent performance measure.

Linear Discriminant Analysis is a widely used classification technique, which is often enhanced by LW-Shrinkage. 
The supplementary material showed that in  LDA the locations of the sample means, relative to the orientations of the covariance, are decisive. They have a huge influence, for example, on the optimal shrinkage intensity w.r.t.\ to classification accuracy.
Optimizing a proxy like ESE does not incorporate information on the location of the class means: the covariance estimates of LW-Shrinkage, NLS and CVC are location-invariant.

LW-Shrinkage is only optimal if the discriminative information is in the high-variance region of the covariance.
The squared error objective, in combination with the limited flexibility of shrinkage, leads to large relative errors in the low-variance part of the spectrum. 

CVC, on the other hand, is more flexible and achieves a good fit for both  small and large eigenvalues. The simulations showed that (i) there is no advantage in using an estimator of the precision oracle and (ii) as the performance of CVC is close to the performance of the oracle, there is no room for improvement by NLS.

\newpage

%\small
\bibliography{ida,machineLearning,finance,dbML}{}